\documentclass{article}

\usepackage[preprint]{corl_2026}
\usepackage{amsmath}
\usepackage{amssymb}
\usepackage{booktabs}
\usepackage{tabularx}
\usepackage{microtype}
\usepackage{graphicx} 

\title{Learning from an Imitable Teacher: Competition Solution for the MetaFine Challenge}

\author{
  Team Friday\\
  \url{https://github.com/YanzeZhang97/MetaFine}
}

\begin{document}
\maketitle

\begin{abstract}
The MetaFine Open Competition (RoboFineMani workshop, CoRL 2026) evaluates fine-grained manipulation beyond binary
success. Its five simulated tasks, from grasping a bottle by an instructed part to inserting letter-shaped pegs
with 4 mm clearance, are scored on perception under lighting and camera changes, instruction understanding, and
behavior. This report presents Team Friday's solution, which got second place with an overall score of 0.947, more
than twice the $\pi_0$ baseline. Instead of fine-tuning a large vision-language-action
(VLA) model, we train one 19M-parameter policy per task with a separate instruction classifier and focus on the
data, all generated in the simulator by a privileged scripted teacher, including its corrections of the student's
mistakes. By analyzing failed episodes, we find that a teacher that almost always succeeds can still teach a
memoryless student badly: its pauses become stalls, its fixed habits become shortcuts, and it never shows the
errors the student makes. Making the teacher \emph{imitable} raised the official score on letter insertion, the
hardest task, from 0.44 to 0.82 with 32\% more data and the same architecture, and finally to 0.88. We summarize
practical lessons on demonstration data and evaluation, which we hope will help in imitation learning for precise manipulation.

\end{abstract}

\keywords{Imitation learning, Fine-grained manipulation, VLA models}

\section{Introduction}
\vspace{-0.2cm}
Vision-language-action (VLA) models~\citep{brohan2023rt,kim2024openvla,black2024pi_0} are large pretrained policies
that follow varied instructions and can be fine-tuned to new tasks, but precision remains hard: the 2025 BEHAVIOR
Challenge winner, a fine-tuned VLA, reports clumsy manipulation as its dominant failure~\citep{larchenko2025task}.

Precise imitation is sensitive to the demonstrations: errors compound into unseen states~\citep{ross2011reduction}, and
policies may copy spurious cues~\citep{de2019causal,geirhos2020shortcut}. Noise injection~\citep{laskey2017dart}
and interactive corrections~\citep{kelly2019hg} widen the states covered, yet successful demonstrations still
differ in how well a policy learns from them~\citep{mandlekar2021matters,belkhale2023data}.

The MetaFine Open Competition provides a diagnostic benchmark for this problem along the three dimensions of MetaFine~\citep{xu2026beyond}: understanding, perception and controlled
behavior~\citep{xu2026beyond}. In a
ManiSkill3-based simulator~\citep{tao2025demonstrating}, a Franka arm with a base and a hand camera follows instructions
to grasp a bottle by its cap or body (T1), move a mug left, right or forward (T2), toggle the red or blue switch (T3),
put one cube into the left box and the rest into the right (T4), or insert a letter peg (C, o, R, L) into its slot
with about 4\,mm of clearance (T5). A task's score averages Perception (P: area under the success curve over camera
and lighting perturbations), Understanding (U: success rate over 20 fixed seeds per instruction variant) and Behavior
(B: success or stage progress). The organizers' $\pi_0$ baseline~\citep{black2024pi_0}, fine-tuned per task on
200--400 planner demonstrations, scores 0.430 (Table~\ref{tab:main}).

In this report, we examine how far small task-specific policies (Sec.~\ref{sec:arch}) can go when the effort goes
into the data. All training data come from a privileged scripted teacher and its corrections of the student
(Sec.~\ref{sec:data}). We made this teacher imitable, since one that always succeeds can still teach a memoryless
student badly through pauses, fixed habits and abrupt moves. This took T5 from 0.44 to 0.88 and T3 from 0.92 to 1.00
officially, and we placed second among 16 teams with 0.947 (Sec.~\ref{sec:exp}).

\section{Architecture}
\vspace{-0.2cm}
\label{sec:arch}
\begin{figure}[t]
    \centering
    \includegraphics[width=0.9\linewidth]{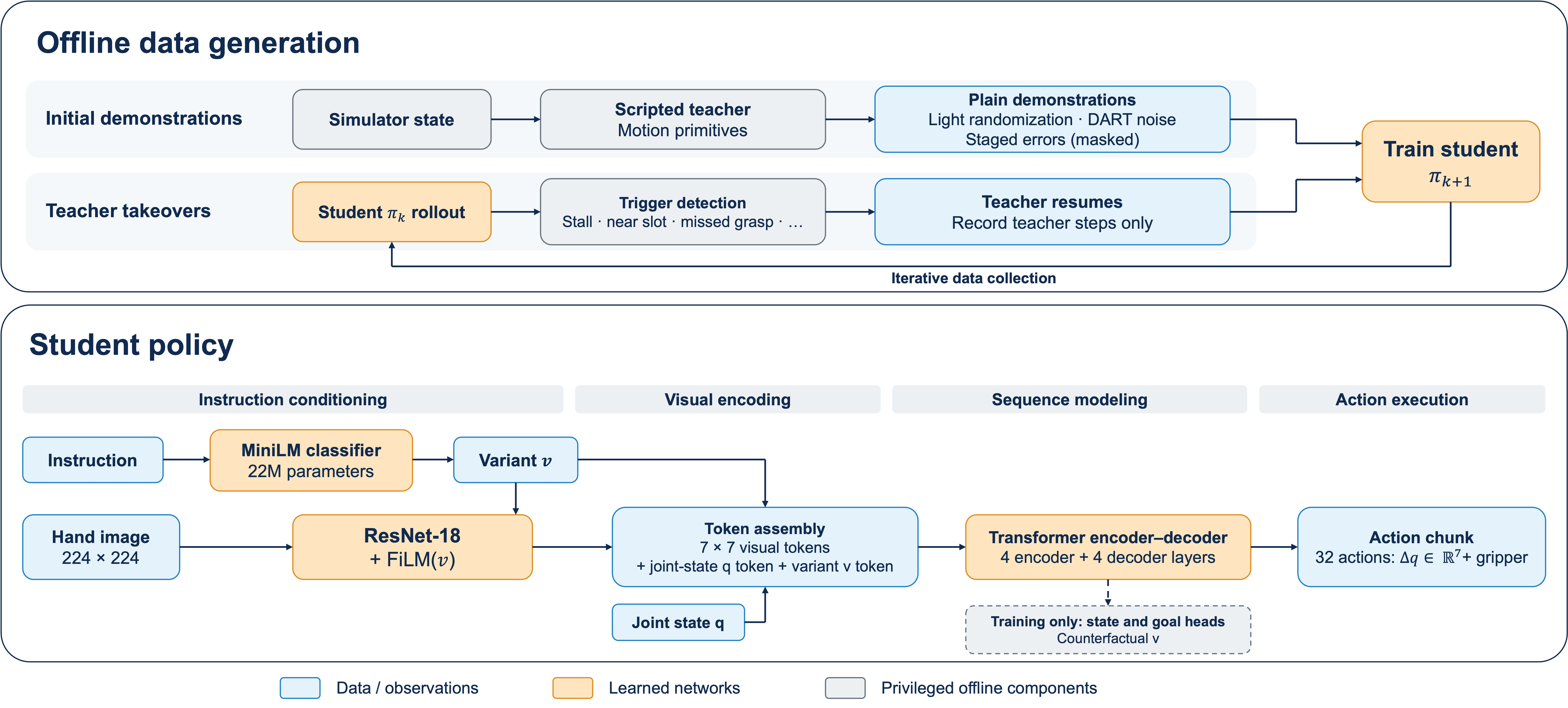}
    \caption{Overview. Top: offline data generation with a privileged teacher and takeovers (Sec.~\ref{sec:data}). 
Bottom: the student policy (Sec.~\ref{sec:arch}), one control network per task plus a shared instruction 
classifier. For T5, a second image stream sees a gripper crop at about twice the resolution.}
\label{fig:overview}
\vspace{-0.5cm}
\end{figure}

As shown in Fig.~\ref{fig:overview} (bottom), our system separates reading the instruction from acting on it. A
sentence encoder \textbf{Instruction classifier} (all-MiniLM-L6-v2~\citep{wang2020minilm, reimers2019sentence}) with a linear head per task maps the
instruction to a variant index, so the control networks see no language. Each task has one action-chunking
transformer (ACT)~\citep{zhao2023learning} (\textbf{Visuomotor policy}) without the CVAE, following the standard design: an ImageNet-initialized
ResNet-18~\citep{he2016deep}, with FiLM~\citep{perez2018film} on the variant, turns the hand image into $7{\times}7$
tokens, and a transformer (4+4 layers, $d=256$) maps them, with joint-position and variant tokens, to 32 actions
(7 joint deltas, gripper), executing 8. Fine-tuned on paraphrases from a small grammar per task with negations
(``not the cap''), distractors, attribute binding and typos, the classifier reads all 14 official instructions and
226 held-out paraphrases correctly.

For training, two extra heads (dashed box in Fig.~\ref{fig:overview}) help the policy use the variant: one regresses
object states, and the other predicts the instructed object's position and feeds the decoder. For 30\% of the samples
we feed another variant code and supervise only this goal head, with that variant's object position. Without the goal
head, our T3 policy ignored the variant (blue episodes pushed the red switch) and solved 30 of 40 clean episodes,
against 78 of 80 with it on the same data.

\section{Dataset}
\label{sec:data}
\vspace{-0.2cm}
\subsection{Teacher and Student Model}
\vspace{-0.2cm}
Consider a memoryless student that observes $o_t$, the hand image, the joint positions and the variant, and executes
8 of its 32 predicted actions before observing again (Sec.~\ref{sec:arch}). Its training data come from a privileged
teacher that runs scripted routines (motion primitives with inverse kinematics) with perception answered from the
simulator state $s_t$, keeping its survey pose and close looks so that the target stays in the hand camera's view, a requirement shared with
image-based visual servoing~\citep{zhang2023occlusion}.
The
teacher provides two kinds of data. Plain demonstrations randomize the ambient light, inject DART
noise~\citep{laskey2017dart} into the executed actions while recording clean labels, continue briefly after success so
that the student learns to stay put, and cover every seed under every variant. The teacher can also stage an error,
such as a misaligned descent, whose frames are masked from the loss, followed by a labelled recovery. Takeovers follow
HG-DAgger~\citep{kelly2019hg, ross2011reduction}: a trained student drives until a trigger fires (it stalls, closes
on nothing, nears the board, \ldots), after which the teacher resumes and only its steps are recorded. We keep only
successful episodes, and training seeds are disjoint from evaluation seeds.

\begin{table}[t]
\centering
\caption{Official results (task score: mean of P, U, B; overall: mean over tasks). Episodes: training episodes per
task (ours: teacher demonstrations and takeovers; $\pi_0$: planner demonstrations).}
\label{tab:main}
\small
\setlength{\tabcolsep}{4.2pt}
\begin{tabular}{l rcccc c rcccc}
\toprule
 & \multicolumn{5}{c}{Ours (Team Friday)} & & \multicolumn{5}{c}{$\pi_0$ baseline} \\
\cmidrule{2-6}\cmidrule{8-12}
Task & Episodes & P & U & B & Score & & Episodes & P & U & B & Score \\
\midrule
T1 grasp part & 1,410 & 0.865 & 0.900 & 0.900 & 0.888 & & 200 & 0.296 & 0.35 & 0.35 & 0.332 \\
T2 move mug & 1,192 & 0.983 & 0.983 & 0.983 & 0.983 & & 300 & 0.892 & 0.95 & 0.95 & 0.931 \\
T3 toggle switch & 1,955 & 1.000 & 1.000 & 1.000 & 1.000 & & 200 & 0.346 & 0.60 & 0.60 & 0.515 \\
T4 cubes into boxes & 896 & 0.995 & 0.983 & 0.983 & 0.987 & & 300 & 0.192 & 0.35 & 0.45 & 0.331 \\
T5 insert letter & 9,674 & 0.880 & 0.875 & 0.875 & 0.877 & & 400 & 0.025 & 0.05 & 0.05 & 0.042 \\
\midrule
Overall & & & & & \textbf{0.947} & & & & & & 0.430 \\
\bottomrule
\end{tabular}
\end{table}

\begin{table}[t]
\centering
\caption{T5 development. Local: successes in 80 clean episodes (same seeds in every row). Official: leaderboard T5
score. The network is unchanged from t5\_c on (t5\_a: centre instead of gripper crop, fewer auxiliary targets);
t5\_j was fine-tuned from t5\_g, the others trained from scratch.}
\label{tab:t5}
\small
\begin{tabularx}{\textwidth}{l r X c c}
\toprule
Network & Episodes & Change in the data & Local & Official \\
\midrule
t5\_a & 1,541 & plain teacher demonstrations & 13/80 & -- \\
t5\_c & 4,067 & off-centre grasps, seating from the true in-hand pose, staged misalignments; takeovers & 34/80 & 0.442 \\
t5\_f & 3,818 & settles capped at 3 steps, release once the peg is home, descent noise; takeovers & 61/80 & -- \\
t5\_g & 5,358 & takeovers resume from the current height instead of rising first & 71/80 & 0.817 \\
t5\_j & 9,674 & release only once seated; two more takeover rounds & -- & 0.875 \\
\bottomrule
\end{tabularx}
\end{table}

\subsection{Making the Teacher Imitable}
\label{sec:imitable}
\vspace{-0.2cm}
Success alone does not make a teacher imitable. We call the teacher \emph{imitable} if (i) its action is a stable
function of $o_t$, so that states the student cannot tell apart receive the same action, and (ii) its margins exceed
the student's errors, so that its behavior still succeeds from the states the student reaches. We traced each
student failure to a violated condition and fixed it in the teacher, never at run time.

\textbf{Pauses become stalls.} The teacher's settling waits lasted up to 13--25 steps, longer than the student's
8-step horizon. A student stopped in such a state sees a frame labelled ``hold'', predicts ``hold'' again and never
leaves, which violates condition~(i). Action chunking was introduced partly to cope with such
pauses~\citep{zhao2023learning}, and single-step policies are known to get stuck on idle actions~\citep{chi2025diffusion}.
Here, chunking did not prevent it. One T5 network scored 0 of 240, typically holding the seated peg until the time
limit, as the teacher held a seated peg still for 10--15 steps before opening. Capping waits at 3 steps,
releasing as soon as the peg is home and closing the fingers while moving removed the stalls, which also caused
about a third of the T3 failures and 10 of 400 T1 episodes.

\textbf{Fixed habits become shortcuts.} The T3 teacher always began its push 4\,cm above the contact point. On the
teacher's own states, a 4\,cm stroke and a push to the switch are indistinguishable, so the student learned the
stroke, a case of causal confusion~\citep{de2019causal} or shortcut learning~\citep{geirhos2020shortcut}. It then
turned a 10\,mm hover error into an 11\,mm push error and missed the official contact gate, which the teacher itself
cleared by only 4\,mm (condition~(ii)). Drawing the pre-push distance from 2.5--7\,cm, with up to 4\,mm of lateral
jitter, made the student read the push height from the image. Together with capped waits, this raised T3 from 439
to 497 of 500 paired local episodes (McNemar $p<10^{-4}$) and from 0.919 to 1.000 officially, although the new data
set was also twice as large.

\textbf{Unseen errors leave no recovery.} The T5 teacher grasped exactly where it planned and never showed a peg held
off-centre, whereas the student's grasps landed a few millimetres along the stroke, so it pressed the peg 6--13\,mm off
a slot with 4\,mm of clearance (condition~(ii)). The revised teacher jitters its grasp by up to 6\,mm, seats the peg
from its true in-hand pose and stages misaligned descents with recoveries, and takeovers add the errors the student
actually makes. These corrections must agree with the plain demonstrations (condition~(i)): our takeovers first rose
2\,cm above the grasp, whereas demonstrations descend directly, and t5\_f hovered 1.5--2\,cm above the grasp, where the
two balance, in 48 of its 83 failures on fresh seeds. Resuming from the current height, with one more takeover round,
cut never-grasped failures from 88 to 11 in 480 fresh episodes.

\textbf{Abrupt transitions amplify errors.} The teacher opened the gripper at once, but the student released a step
early and slightly off, so the jolt turned the o by 12--15$^\circ$, beyond the 10$^\circ$ tolerance
(condition~(ii)). Opening over 8 steps raised local o success from 0.81 to 0.92, but not the official score
(Sec.~\ref{sec:eval}).

\section{Experiments}
\label{sec:exp}
\vspace{-0.2cm}
\subsection{Implementation Details}
\vspace{-0.2cm}
We train all networks with AdamW (learning rate $10^{-4}$, $10^{-5}$ for the backbone) and batch size 64. Arm
actions use an L1 loss and the gripper cross-entropy, except in T5, where the teacher pre-shapes the fingers
and the gripper is regressed with L1. We augment images with random shifts of up to 8\,px~\citep{kostrikov2020image}
and use weight averaging. At test time only the two networks run on the allowed
observations. The teacher, simulator state and motion primitives serve only to generate data, and the only external
weights are ImageNet and MiniLM initializations.

\subsection{Main Results}
\vspace{-0.2cm}
\label{sec:eval}

Table~\ref{tab:main} compares our official results with the organizers' $\pi_0$ baseline. The gap grows with the
precision a task requires: on T2, moving a mug, the baseline already reaches 0.93, whereas we raise T4 from 0.33 to
0.99 and T5, the hardest task, from 0.04 to 0.88. Table~\ref{tab:t5} traces how T5 got there. The first network,
trained on 1,541 plain demonstrations, solved only 13 of 80 local episodes. Each later version changed what the
teacher showed (Sec.~\ref{sec:imitable}) and added takeovers from the latest student, and the gains followed these
changes rather than the amount of data: t5\_f was trained on 6\% fewer episodes than t5\_c yet solved 61 instead of
34, and from t5\_c to t5\_g the official score nearly doubled (0.442 to 0.817) with 32\% more episodes and the same
network.

Local and official numbers differ, however. Official scores were typically 0.03--0.08 below our local
estimates, and with 20 fixed seeds per variant one T1 episode moves the overall score by 0.003; even identical T1
weights scored 0.894 and then 0.888, as T1 rendering is not bit-reproducible. Two changes that won paired local
tests, a T1 network (45 vs.\ 19 discordant episodes, $p=0.002$) and the T5 release fix (69 vs.\ 47, $p=0.05$),
scored slightly lower officially (0.867 vs.\ 0.894 and 0.869 vs.\ 0.875), so our final submission keeps the earlier
networks.

\section{Conclusion}
\label{sec:conclusion}

In this report, we presented our solution to the MetaFine Open Competition: one small policy per task, trained
entirely on data from a privileged scripted teacher. The difficulty lay in millimetre-scale execution rather than in
language or perception, and it responded to the data rather than the network: making the teacher imitable raised T5
from 0.44 to 0.88 and T3 from 0.92 to 1.00 officially. We believe that how imitable demonstrations are deserves as much attention as how many there are, and we hope our
findings help others audit their demonstrations, collect consistent corrections and compare versions on paired
seeds~\citep{kress2024robot,agarwal2021deep}.

\textbf{Limitations.}
T1 and T5 remain the weakest tasks, the teacher needs privileged simulator state,
several fixes were introduced together, and all results are in
simulation. We also found each imitability problem by hand; detecting such problems automatically would make the
process systematic. Our takeover triggers are also hand-written. Uncertainty estimates, for example from adaptive
conformal prediction~\citep{zhou2024safety, zhou2025computationally}, could decide when the teacher takes over, as in risk-gated
interactive imitation learning~\citep{hoque2021thriftydagger}.

\subsubsection*{Authors}
Yanze Zhang, Hao Liu, Haochang Hao, Zhixiang Shen, Chunyu Miao, Yuyao Yang

\acknowledgments{We thank the MetaFine organizers for running the competition and for answering our questions.}

\bibliography{refs}

\end{document}